\documentclass{article}

\usepackage{ijcai17}
\usepackage{enumerate}

\usepackage{times}
\usepackage{color}
\usepackage{booktabs}
\usepackage{amsmath}
\usepackage{graphicx}
\usepackage{subfigure} 
\usepackage{multirow}
\usepackage{multicol}
\usepackage{amsthm}
\usepackage[ruled,vlined,shortend,linesnumbered]{algorithm2e} % algorithm
\usepackage[]{algorithmic} % algorithm
\usepackage[hidelinks]{hyperref}
\usepackage{balance}
\usepackage{paralist}

\usepackage{pgf,tikz}
\usepackage{mathrsfs}
\usepackage{fix-cm} 
\usepackage{xcolor}
\usetikzlibrary{arrows.meta}

\newtheorem{definition}{Definition}
\newtheorem{theorem}{Theorem}
\newtheorem{claim}{Claim}

\newcommand{\hide}[1]{}

\newcommand{\bit}{\begin{compactitem}}
\newcommand{\eit}{\end{compactitem}}
\newcommand{\ben}{\begin{compactenum}}
\newcommand{\een}{\end{compactenum}}

\newcommand{\SA}{\mathcal{A}}
\newcommand{\SG}{\mathcal{G}}
\newcommand{\SV}{\mathcal{V}}
\newcommand{\SE}{\mathcal{E}}
\newcommand{\SSS}{\mathcal{S}}
\newcommand{\SF}{\mathcal{F}}
\newcommand{\SN}{\mathcal{N}}
\newcommand{\ST}{\mathcal{T}}
\newcommand{\SC}{\mathcal{C}}
\newcommand{\SB}{\mathcal{B}}
\newcommand{\kneighbor}{\SN_i(k)}

\newcommand{\POANE}{\mathsf{PoA}}
\newcommand{\POSNE}{\mathsf{PoS}}

\newcommand{\change}{}

\makeatletter
\newcommand\HUGE{\@setfontsize\Huge{36}{38}}
\makeatother   

\title{Why You Should Charge Your Friends for Borrowing Your Stuff}
\author{Kijung Shin,\ \  Euiwoong Lee,\ \  Dhivya Eswaran,\ \ and Ariel D. Procaccia\\ 
	Carnegie Mellon University, Pittsburgh, PA, USA\\
	\{kijungs, euiwoonl, deswaran, arielpro\}@cs.cmu.edu}

\begin{document}

\maketitle

\begin{abstract}
We consider goods that can be shared with $k$-hop neighbors (i.e., the set of nodes within $k$ hops from an owner) on a \change{social} network.
%\change{If people can just borrow the good from their peers in the network, then they might not have any incentive to buy it.
%Therefore, in this work,} 
We examine incentives to buy such a good by devising game-theoretic models where each node decides whether to buy the good or free ride.
First, we find that social inefficiency, specifically \change{excessive purchase} of the good, occurs in Nash equilibria.
Second, the social inefficiency decreases as $k$ increases and thus a good can be shared with more nodes.
Third, and most importantly, the social inefficiency can also be significantly reduced by \change{charging free riders an access cost and paying it to owners}, leading to the conclusion that organizations and system designers should impose such a cost.
These findings are supported by our theoretical analysis in terms of the price of anarchy and the price of stability; and by simulations based on synthetic and real 
\change{social} networks. 

%Our work considers a good that can be shared with $k$-hop neighbors (i.e., set of nodes within $k$ hops from a buyer) on a social network.
%We propose a game-theoretic model where each node decides whether to buy the good or free ride.
%We find that social inefficiency, specifically overproduction of the good, occurs in Nash equilibria.
%However, the social inefficiency can be significantly reduced by imposing a cost to free riders.
%We show these by (a) theoretical analysis using the price of anarchy and the price of stability; and (b) simulation using best response dynamics.

%Our work considers public goods which are non-excludable along edges in a social or geographic networks.
%We propose a game-theoretic model where each node decides whether to buy a good or free ride.
%The simplicity of our model, as compared to previous work, allows us to find socially optimal outcomes by solving an integer programming problem. However, players act independently, resulting in a Nash equilibria which are significantly worse than the social optimum. 
%Our work identifies the reason for this gap and addresses this by (a) central planning using structural properties of the network, (b) imposing proper cost on free riders.
%Through experiments on various synthetic and real world networks, we demonstrate that the proposed techniques result in up to 111 times better outcome.
\end{abstract}

\section{Introduction}
\label{sec:intro}
Social networks are known to play an important role in the everyday choices people make.
In particular, a significant body of work studies the {\it network effect}, %in which people incur an explicit benefit when they and their neighbors buy compatible goods instead of incompatible ones.
in which there are payoffs from aligning one's decision with those of others \cite{markus1987toward,blume1993statistical,ellison1993learning,rogers2010diffusion}.
For example, direct payoffs arise when friends or collaborators use compatible technologies instead of incompatible ones, that is, the game rewards \emph{coordination}. 

Our work considers the purchase of \emph{shareable goods}, which, in a sense, gives rise to a certain type of \emph{anti-coordination} game.
Indeed, buying such a good yields a benefit only when no friend buys the good, since otherwise free riding is possible. An example that would be familiar to most parents is seldom-used baby gear, such as portable cribs, which are frequently borrowed by friends or friends of friends; similar examples include ski gear and hiking equipment. Expensive lab equipment provides a more pertinent example: Confocal laser scanning microscopes, or polymerase chain reaction (PCR) machines, are typically bought by one investigator and used by collaborators. In the realm of AI, one can imagine a multi-agent system populated by heterogeneous software agents that interact and share special computational resources, e.g., a high-end graphics processing unit (GPU) for particularly demanding image processing tasks.

%For example, borrowing camping equipments, instead of buying them, saves expense if you have friends who bought them.
%Likewise, seeking advice, instead of doing research, saves effort if you have friends who did the same research before.
%%For example, installing a pollution abatement system in a house benefits the neighbors as well, friends and colleagues utilize one's innovations/ideas for their personal purposes.

To examine incentives to buy shareable goods, we devise game-theoretic models where each node on a network decides whether to buy a good that is shareable with {\it $k$-hop neighbors} (i.e., nodes within $k$ hops from an owner), or free ride.
Specifically, the good in question is {\it non-excludable} and {\it non-rivalrous} in that no $k$-hop neighbor can be excluded from use, and use by a neighbor does not reduce availability to others. Note that the goods in the examples given above are (essentially) non-rivalrous, as any single person (or agent) requires the good only from time to time. 

We find that social inefficiency, specifically \change{excessive purchase} of the good, occurs in \change{Nash equilibria}. 
Moreover, the social inefficiency decreases as $k$ increases and thus a good is shared with more people.
Finally, \change{charging free riders an access cost and paying it to owners} also significantly reduces the social inefficiency.
We support these findings both theoretically and experimentally.

%imposing an access cost to free riders significantly reduces the social inefficiency. Finally, the social inefficiency is minimized with lower access cost as $k$ increases.

Compared with previous work on shareable goods on a network, discussed in Section~\ref{sec:related}, our contributions are as follows:
\bit
	\item \emph{Efficiency Analysis of Equilibria :} We provide worst-case analysis of the efficiency of equilibria, in terms of the price of anarchy and the price of stability. %applies to any general network not only for toy networks such as star or complete graphs.
	\item \emph{Simulation on Real-world Networks:} The simplicity of our model allows us to measure social inefficiency on real-world social networks, with thousands of nodes, through simulations of best-response dynamics.
	\item \emph{Mechanism Design:} We analyze the effects of access costs on social inefficiency and suggest an appropriate cost for minimizing social inefficiency. 
\eit

The rest of the paper is organized as follows. 
In Section~\ref{sec:game}, we define shareable goods games on a network.
In Section~\ref{sec:analysis}, we give a theoretical analysis of the efficiency of equilibria.
In Section~\ref{sec:simulation}, we present simulation results.
After discussing related work in Section~\ref{sec:related}, we draw conclusions in Section~\ref{sec:conclusion}.

\section{Our Models}
\label{sec:game}
In this section, we formally define two game-theoretic models of the purchase of shareable goods on a network.
%We first present the basic shareable goods game (SGG), then extend it to the shareable goods game with access costs (SGG-AC).

\subsection{Shareable Goods Game (SGG)}
\label{sec:game:basic}

Consider an undirected network $\SG = (\SV,\SE)$ where $\SV=\{1,2,...,n\}.$ 
The \emph{players} of the game are nodes in $\SG $.
Each node decides whether to buy a good or not.
The \emph{strategy} of node $i$ is denoted by $s_{i}\in \SSS_{i}$, where $\SSS_{i}=\{0,1\}$ denotes the \emph{strategy set} of node $i$.
If node $i$ buys the good then $s_{i}=1$, and otherwise $s_{i}=0$ (only pure strategies are considered).
Given any \emph{strategy profile} $s =
(s_{1}, s_{2},..., s_{n})$, we use $s_{-i}$ to denote the strategies taken by all nodes but $i$. 
Then, $s$ is also denoted by $(s_{i}, s_{-i})$.
The \emph{price} of a good is $p$ ($>0$), which is identical for all nodes.
A node gets \emph{benefit} $b$ ($>p$) by having access to a good and $0$ otherwise. 
Each node $i$ can access a good if it buys the good itself or has at least one node who buys the good within $k$ \change{$(\geq 1)$} hops.
We assume that having access to multiple goods does not increase the benefit of a node and that being accessed by multiple nodes does not decrease the benefit derived from a good ({\it non-rivalry}).

\begin{table}[t]
	\vspace{-3mm}
	\centering
	\caption{\label{tab:utility} Utility in an SGG. }
	{\renewcommand{\arraystretch}{1.2}
		\begin{tabular}{c|c|c}
			\toprule
			state & conditions \ & utility (i.e., $u_{i}$) \\
			\midrule
			\textbf{buy} & $s_{i}=1$ & $b-p$ ($>0$)\\
			\textbf{free ride} & $s_{i}=0,$ $\sum_{j\in \kneighbor}s_{j}\geq 1$& $b$ ($>0$) \\
			\textbf{no access} & $s_{i}=0,$ $\sum_{j\in \kneighbor}s_{j}=0$ & $0$ \\
			\bottomrule
		\end{tabular}
	}
\end{table}
\begin{table}[t]
	\vspace{-3mm}
	\centering
	\caption{Utility in an SGG-AC.}
	{\renewcommand{\arraystretch}{1.2}
		\begin{tabular}{c|c|c}
			\toprule
			state & conditions \ & utility (i.e., $u_{i}$) \\
			\midrule
			\textbf{buy} & $s_{i}=i$ & $b-p+a|\SF_{i}| (> 0)$ \\
			\change{\textbf{rent}} & $s_{i}=j(\neq i)$, $s_{j}=j$ & $b-a (> 0)$ \\
			\textbf{no access} & $s_{i}=j(\neq i)$, $s_{j}\neq j$ & $0$ \\
			\bottomrule
		\end{tabular}
	}
	\label{tab:utility:cost}
\end{table}

In this setting, the \emph{utility} $u_{i}(s)$ of node $i$ under strategy profile $s$ depends on the strategies of its \emph{$k$-hop neighbors} $\kneighbor$ (i.e., the set of nodes within $k$-hops from $i$ including $i$ itself), as given in Table \ref{tab:utility}.
Note that each node gets the highest utility $b$ when it free rides and the second highest utility $b-p$ when it buys the good.
Each node gets the lowest utility $0$ when neither the node nor its $k$-hop neighbors buy the good. 
%The \textbf{social utility} $U=\sum_{i \in \SV}u_i$ of an outcome of the game would then be the sum of the utilities of all the nodes in the network.
%If we let \textbf{social benefit} $U_{b}$ be the sum of benefits of all players, and \textbf{social cost} be the total cost used to buy goods (i.e., $U_{c}=p\sum_{i=1}^{|\SV|}s_{i}$),
%social utility is the difference between social benefit and social cost (i.e., $U=U_b-U_c$).

%Define the game and basic terms (e.g., social cost, eash Equilibria, and strong Nash equilibria)
%Give conditions for Nash Equilibria, Strong Nash Equilibria and Social optima
%Give an example of social inefficiency, and explain the need of access costs.
\change{SGG extends the best-shot game \cite{hirshleifer1983weakest}, which is equivalent to SGG if $k=1$, by considering not only direct but $k$-hop neighbors. SGG-AC, discussed in the following subsection, further extends SGG by considering access costs.}

\subsection{Shareable Goods Game with Access Costs (SGG-AC)}
\label{sec:game:cost}

In this subsection, we extend the game defined in the previous section to a game we call the shareable goods game with access costs (SGG-AC), where each free rider has to pay an access cost.
We focus on the differences from an SGG.
 
The \emph{strategy set} of each node $i$ is $\SSS_{i}=\kneighbor$, its $k$-hop neighbors \change{including $i$ itself}.
If node $i$ buys a good then $s_{i}=i$, and if node $i$ does not buy a good but wants to access a good bought by node $j\neq i$ then $s_{i}=j$.
If $s_{i}=j$ for $j\neq i$, and node $j$ actually buys a good (i.e., $s_{j}=j$) then node $i$ derives benefit from the good at the expense of paying an \emph{access cost} of $a$ ($< p$) to node $j$.
The \emph{followers} of node $i$, the set of nodes who want to access the good bought by node $i$, are denoted by $\SF_{i}=\{j\in \kneighbor\backslash\{i\}: s_{j}=i\}$.
Then,  the \emph{utility} $u_{i}(s)$ of node $i$ under strategy profile $s$ in an SGG-AC is given in Table~\ref{tab:utility:cost}.
Define the \emph{follower threshold} $\xi=\lceil p/a \rceil -1$; for ease of exposition we assume that $p/a$ is not an integer. If node $i$ has at least $\xi$ followers (i.e., $|\SF_{i}|\geq \xi$), it has the highest utility when it buys a good (i.e., $s_{i}=i$). Otherwise, \change{renting a good} is preferred.
Each node has the lowest utility when it is not accessing any good.

\section{Analysis of Equilibria}
\label{sec:analysis}
\begin{figure}[t]
	\centering
	\vspace{-3mm}
	\begin{tikzpicture}[line cap=round,line join=round,x=0.72cm,y=0.72cm]
	\clip(-1.62,4.5) rectangle (2.78,5.4);
	\draw (0.75,5.4) node[anchor=north west] {{Not Buy}};
	\draw (-1.15,5.4) node[anchor=north west] {{Buy}};
	\draw [color=black, fill=black,line width=2.pt] (-1.34906,5.1) circle (4pt);
	\draw [color=black, fill=white,line width=2.pt] (0.54172,5.1) circle (4pt);
	\end{tikzpicture}\\
	\vspace{-2mm}
	\subfigure[Not NE]{ \label{fig:game:basic:no_ne}
		\begin{tikzpicture}[line cap=round,line join=round,x=0.64cm,y=0.64cm]
		\clip(-1.6,0.5) rectangle (1.6,3.53);
		\fill[color=white,fill=white,fill opacity=0] (-0.8,0.8) -- (0.8,0.8) -- (1.2944271909999159,2.3216904260722453) -- (0.,3.2621468297402023) -- (-1.2944271909999157,2.321690426072246) -- cycle;
		\draw [line width=2.pt] (-0.8,0.8)-- (0.8,0.8);
		\draw [color=white] (0.8,0.8)-- (1.2944271909999159,2.3216904260722453);
		\draw [color=white] (1.2944271909999159,2.3216904260722453)-- (0.,3.2621468297402023);
		\draw [color=white] (0.,3.2621468297402023)-- (-1.2944271909999157,2.321690426072246);
		\draw [color=white] (-1.2944271909999157,2.321690426072246)-- (-0.8,0.8);
		\draw [line width=2.pt] (-1.2944271909999157,2.321690426072246)-- (0.,1.90110553638);
		\draw [line width=2.pt] (0.,3.2621468297402023)-- (0.,1.90110553638);
		\draw [line width=2.pt] (0.,1.90110553638)-- (1.2944271909999159,2.3216904260722453);
		\draw [line width=2.pt] (0.,1.90110553638)-- (0.8,0.8);
		\draw [line width=2.pt] (0.,1.90110553638)-- (-0.8,0.8);
		\draw [line width=2.pt] (1.2944271909999159,2.3216904260722453)-- (0.8,0.8);
		\draw [line width=2.pt] (-0.8,0.8)-- (0.8,0.8);
		\begin{normalsize}
		\draw [color=black ,fill=white,line width=2.pt] (-0.8,0.8) circle (4pt);
		\draw[color=black,fill=white] (-1.0682536183519724,1.2) node {5};
		\draw [color=black, fill=black,line width=2.pt] (0.8,0.8) circle (4pt);
		\draw[color=black,fill=white] (1.2,1.2) node {6};
		\draw[color=white] (0.6353240941552338,1.912075524355764) node {};
		\draw [color=black,fill=white,line width=2.pt] (1.2944271909999159,2.3216904260722453) circle (4pt);
		\draw[color=black ] (1.2944271909999159,2.8) node {4};
		\draw [color=black ,fill=white,line width=2.pt] (0.,3.2621468297402023) circle (4pt);
		\draw[color=black] (0.45,3.3) node {1};
		\draw [color=black ,fill=white,line width=2.pt] (-1.2944271909999157,2.321690426072246) circle (4pt);
		\draw[color=black] (-1.2944271909999157,2.8) node {2};
		\draw [color=black, fill=black,line width=2.pt] (0.,1.9) circle (4pt);
		\draw[color=black] (0.3,2.4) node {3};
		\end{normalsize}
		\end{tikzpicture}
	}
	\subfigure[NE (\change{$cost=4p$})]{ \label{fig:game:basic:ne:ineff}
		\ \ \
		\begin{tikzpicture}[line cap=round,line join=round,x=0.64cm,y=0.64cm]
		\clip(-1.6,0.5) rectangle (1.6,3.53);
		\fill[color=white,fill=white,fill opacity=0] (-0.8,0.8) -- (0.8,0.8) -- (1.2944271909999159,2.3216904260722453) -- (0.,3.2621468297402023) -- (-1.2944271909999157,2.321690426072246) -- cycle;
		\draw [line width=2.pt] (-0.8,0.8)-- (0.8,0.8);
		\draw [color=white] (0.8,0.8)-- (1.2944271909999159,2.3216904260722453);
		\draw [color=white] (1.2944271909999159,2.3216904260722453)-- (0.,3.2621468297402023);
		\draw [color=white] (0.,3.2621468297402023)-- (-1.2944271909999157,2.321690426072246);
		\draw [color=white] (-1.2944271909999157,2.321690426072246)-- (-0.8,0.8);
		\draw [line width=2.pt] (-1.2944271909999157,2.321690426072246)-- (0.,1.90110553638);
		\draw [line width=2.pt] (0.,3.2621468297402023)-- (0.,1.90110553638);
		\draw [line width=2.pt] (0.,1.90110553638)-- (1.2944271909999159,2.3216904260722453);
		\draw [line width=2.pt] (0.,1.90110553638)-- (0.8,0.8);
		\draw [line width=2.pt] (0.,1.90110553638)-- (-0.8,0.8);
		\draw [line width=2.pt] (1.2944271909999159,2.3216904260722453)-- (0.8,0.8);
		\draw [line width=2.pt] (-0.8,0.8)-- (0.8,0.8);
		\begin{normalsize}
		\draw [color=black ,fill=black,line width=2.pt] (-0.8,0.8) circle (4pt);
		\draw[color=black,fill=white] (-1.0682536183519724,1.2) node {5};
		\draw [color=black, fill=white,line width=2.pt] (0.8,0.8) circle (4pt);
		\draw[color=black,fill=white] (1.2,1.2) node {6};
		\draw[color=white] (0.6353240941552338,1.912075524355764) node {};
		\draw [color=black,fill=black,line width=2.pt] (1.2944271909999159,2.3216904260722453) circle (4pt);
		\draw[color=black ] (1.2944271909999159,2.8) node {4};
		\draw [color=black ,fill=black,line width=2.pt] (0.,3.2621468297402023) circle (4pt);
		\draw[color=black] (0.45,3.3) node {1};
		\draw [color=black ,fill=black,line width=2.pt] (-1.2944271909999157,2.321690426072246) circle (4pt);
		\draw[color=black] (-1.2944271909999157,2.8) node {2};
		\draw [color=black, fill=white,line width=2.pt] (0.,1.9) circle (4pt);
		\draw[color=black] (0.3,2.4) node {3};
		\end{normalsize}
		\end{tikzpicture}
		\ \ \
	}
	\subfigure[NE (\change{$cost=p$})]{ \label{fig:game:basic:ne:eff}
		\ \
		\begin{tikzpicture}[line cap=round,line join=round,x=0.64cm,y=0.64cm]
		\clip(-1.6,0.5) rectangle (1.6,3.53);
		\fill[color=white,fill=white,fill opacity=0] (-0.8,0.8) -- (0.8,0.8) -- (1.2944271909999159,2.3216904260722453) -- (0.,3.2621468297402023) -- (-1.2944271909999157,2.321690426072246) -- cycle;
		\draw [line width=2.pt] (-0.8,0.8)-- (0.8,0.8);
		\draw [color=white] (0.8,0.8)-- (1.2944271909999159,2.3216904260722453);
		\draw [color=white] (1.2944271909999159,2.3216904260722453)-- (0.,3.2621468297402023);
		\draw [color=white] (0.,3.2621468297402023)-- (-1.2944271909999157,2.321690426072246);
		\draw [color=white] (-1.2944271909999157,2.321690426072246)-- (-0.8,0.8);
		\draw [line width=2.pt] (-1.2944271909999157,2.321690426072246)-- (0.,1.90110553638);
		\draw [line width=2.pt] (0.,3.2621468297402023)-- (0.,1.90110553638);
		\draw [line width=2.pt] (0.,1.90110553638)-- (1.2944271909999159,2.3216904260722453);
		\draw [line width=2.pt] (0.,1.90110553638)-- (0.8,0.8);
		\draw [line width=2.pt] (0.,1.90110553638)-- (-0.8,0.8);
		\draw [line width=2.pt] (1.2944271909999159,2.3216904260722453)-- (0.8,0.8);
		\draw [line width=2.pt] (-0.8,0.8)-- (0.8,0.8);
		\begin{normalsize}
		\draw [color=black ,fill=white,line width=2.pt] (-0.8,0.8) circle (4pt);
		\draw[color=black,fill=white] (-1.0682536183519724,1.2) node {5};
		\draw [color=black, fill=white,line width=2.pt] (0.8,0.8) circle (4pt);
		\draw[color=black,fill=white] (1.2,1.2) node {6};
		\draw[color=white] (0.6353240941552338,1.912075524355764) node {};
		\draw [color=black,fill=white,line width=2.pt] (1.2944271909999159,2.3216904260722453) circle (4pt);
		\draw[color=black ] (1.2944271909999159,2.8) node {4};
		\draw [color=black ,fill=white,line width=2.pt] (0.,3.2621468297402023) circle (4pt);
		\draw[color=black] (0.45,3.3) node {1};
		\draw [color=black ,fill=white,line width=2.pt] (-1.2944271909999157,2.321690426072246) circle (4pt);
		\draw[color=black] (-1.2944271909999157,2.8) node {2};
		\draw [color=black, fill=black,line width=2.pt] (0.,1.9) circle (4pt);
		\draw[color=black] (0.3,2.4) node {3};
		\end{normalsize}
		\end{tikzpicture}
		\ \
	}
	\vspace{-3mm}
	\caption{\label{fig:game:basic} Example strategy profiles in an SGG when $k=1$. (a) is not an NE since each of node 3 and node 6 would be better off not buying. \change{Between the NEs, (c) leads to a lower social cost than (b).}
	}
\end{figure}
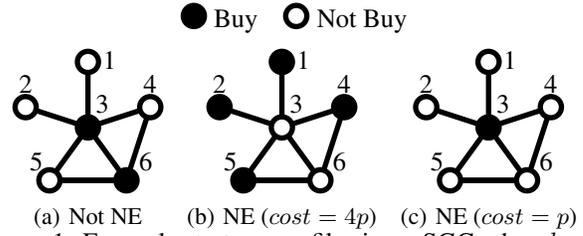

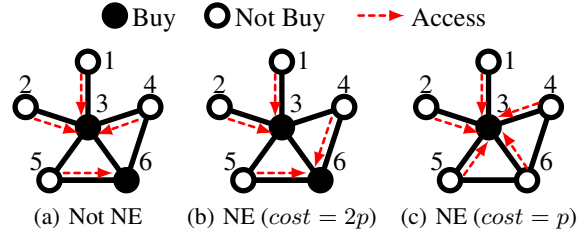
\begin{figure}[t]
	\centering
	\vspace{-3mm}
		\begin{tikzpicture}[line cap=round,line join=round,x=0.72cm,y=0.72cm]
		\clip(-1.63,4.5) rectangle (5.7,5.4);
		\draw (0.7,5.4) node[anchor=north west] {{Not Buy}};
		\draw (-1.2,5.4) node[anchor=north west] {{Buy}};
		\draw (3.95,5.4) node[anchor=north west] {{Access}};
		\draw [color=black, fill=black,line width=2.pt] (-1.34906,5.1) circle (4pt);
		\draw [color=black, fill=white,line width=2.pt] (0.54172,5.1) circle (4pt);
		\draw[-{latex[scale=3.0]},line width=1.pt,dash pattern=on 2pt off 2pt,color=red] (3.15,5.1) -- (4,5.1);
		\end{tikzpicture}\\
	\vspace{-2mm}
	\subfigure[Not NE]{
		\begin{tikzpicture}[line cap=round,line join=round,x=0.64cm,y=0.64cm]
		\clip(-1.6,0.5) rectangle (1.6,3.53);
		\fill[color=white,fill=white,fill opacity=0] (-0.8,0.8) -- (0.8,0.8) -- (1.2944271909999159,2.3216904260722453) -- (0.,3.2621468297402023) -- (-1.2944271909999157,2.321690426072246) -- cycle;
		\draw [line width=2.pt] (-0.8,0.8)-- (0.8,0.8);
		\draw [color=white] (0.8,0.8)-- (1.2944271909999159,2.3216904260722453);
		\draw [color=white] (1.2944271909999159,2.3216904260722453)-- (0.,3.2621468297402023);
		\draw [color=white] (0.,3.2621468297402023)-- (-1.2944271909999157,2.321690426072246);
		\draw [color=white] (-1.2944271909999157,2.321690426072246)-- (-0.8,0.8);
		\draw [line width=2.pt] (-1.2944271909999157,2.321690426072246)-- (0.,1.90110553638);
		\draw [line width=2.pt] (0.,3.2621468297402023)-- (0.,1.90110553638);
		\draw [line width=2.pt] (0.,1.90110553638)-- (1.2944271909999159,2.3216904260722453);
		\draw [line width=2.pt] (0.,1.90110553638)-- (0.8,0.8);
		\draw [line width=2.pt] (0.,1.90110553638)-- (-0.8,0.8);
		\draw [line width=2.pt] (1.2944271909999159,2.3216904260722453)-- (0.8,0.8);
		\draw [line width=2.pt] (-0.8,0.8)-- (0.8,0.8);
		\begin{normalsize}
		\draw [color=black ,fill=white,line width=2.pt] (-0.8,0.8) circle (4pt);
		\draw[color=black,fill=white] (-1.0682536183519724,1.2) node {5};
		\draw [color=black, fill=black,line width=2.pt] (0.8,0.8) circle (4pt);
		\draw[color=black,fill=white] (1.2,1.2) node {6};
		\draw[color=white] (0.6353240941552338,1.912075524355764) node {};
		\draw [color=black,fill=white,line width=2.pt] (1.2944271909999159,2.3216904260722453) circle (4pt);
		\draw[color=black ] (1.2944271909999159,2.8) node {4};
		\draw [color=black ,fill=white,line width=2.pt] (0.,3.2621468297402023) circle (4pt);
		\draw[color=black] (0.45,3.3) node {1};
		\draw [color=black ,fill=white,line width=2.pt] (-1.2944271909999157,2.321690426072246) circle (4pt);
		\draw[color=black] (-1.2944271909999157,2.8) node {2};
		\draw [color=black, fill=black,line width=2.pt] (0.,1.9) circle (4pt);
		\draw[color=black] (0.3,2.4) node {3};
		\draw[-{latex[scale=3.0]},line width=1.pt,dash pattern=on 2pt off 2pt,color=red] (-1.1,2.07)-- (-0.19,1.78);
		\draw[-{latex[scale=3.0]},line width=1.pt,dash pattern=on 2pt off 2pt,color=red] (-0.14,3)-- (-0.14,2.11);
		\draw[-{latex[scale=3.0]},line width=1.pt,dash pattern=on 2pt off 2pt,color=red] (1.03,2.05)-- (0.18,1.75);
		\draw[-{latex[scale=3.0]},line width=1.pt,dash pattern=on 2pt off 2pt,color=red] (-0.5,0.95)-- (0.52,0.95);
		\end{normalsize}
		\end{tikzpicture}
	}
	\subfigure[NE (\change{$cost=2p$})]{
		\ \ \
		\begin{tikzpicture}[line cap=round,line join=round,x=0.64cm,y=0.64cm]
		\clip(-1.6,0.5) rectangle (1.6,3.53);
		\fill[color=white,fill=white,fill opacity=0] (-0.8,0.8) -- (0.8,0.8) -- (1.2944271909999159,2.3216904260722453) -- (0.,3.2621468297402023) -- (-1.2944271909999157,2.321690426072246) -- cycle;
		\draw [line width=2.pt] (-0.8,0.8)-- (0.8,0.8);
		\draw [color=white] (0.8,0.8)-- (1.2944271909999159,2.3216904260722453);
		\draw [color=white] (1.2944271909999159,2.3216904260722453)-- (0.,3.2621468297402023);
		\draw [color=white] (0.,3.2621468297402023)-- (-1.2944271909999157,2.321690426072246);
		\draw [color=white] (-1.2944271909999157,2.321690426072246)-- (-0.8,0.8);
		\draw [line width=2.pt] (-1.2944271909999157,2.321690426072246)-- (0.,1.90110553638);
		\draw [line width=2.pt] (0.,3.2621468297402023)-- (0.,1.90110553638);
		\draw [line width=2.pt] (0.,1.90110553638)-- (1.2944271909999159,2.3216904260722453);
		\draw [line width=2.pt] (0.,1.90110553638)-- (0.8,0.8);
		\draw [line width=2.pt] (0.,1.90110553638)-- (-0.8,0.8);
		\draw [line width=2.pt] (1.2944271909999159,2.3216904260722453)-- (0.8,0.8);
		\draw [line width=2.pt] (-0.8,0.8)-- (0.8,0.8);
		\begin{normalsize}
		\draw [color=black ,fill=white,line width=2.pt] (-0.8,0.8) circle (4pt);
		\draw[color=black,fill=white] (-1.0682536183519724,1.2) node {5};
		\draw [color=black, fill=black,line width=2.pt] (0.8,0.8) circle (4pt);
		\draw[color=black,fill=white] (1.2,1.2) node {6};
		\draw[color=white] (0.6353240941552338,1.912075524355764) node {};
		\draw [color=black,fill=white,line width=2.pt] (1.2944271909999159,2.3216904260722453) circle (4pt);
		\draw[color=black ] (1.2944271909999159,2.8) node {4};
		\draw [color=black ,fill=white,line width=2.pt] (0.,3.2621468297402023) circle (4pt);
		\draw[color=black] (0.45,3.3) node {1};
		\draw [color=black ,fill=white,line width=2.pt] (-1.2944271909999157,2.321690426072246) circle (4pt);
		\draw[color=black] (-1.2944271909999157,2.8) node {2};
		\draw [color=black, fill=black,line width=2.pt] (0.,1.9) circle (4pt);
		\draw[color=black] (0.3,2.4) node {3};
		\draw[-{latex[scale=3.0]},line width=1.pt,dash pattern=on 2pt off 2pt,color=red] (-1.1,2.07)-- (-0.19,1.78);
		\draw[-{latex[scale=3.0]},line width=1.pt,dash pattern=on 2pt off 2pt,color=red] (-0.14,3)-- (-0.14,2.11);
		\draw[-{latex[scale=3.0]},line width=1.pt,dash pattern=on 2pt off 2pt,color=red] (1.03,2.09)-- (0.68,1.06);
		\draw[-{latex[scale=3.0]},line width=1.pt,dash pattern=on 2pt off 2pt,color=red] (-0.5,0.95)-- (0.52,0.95);
		\end{normalsize}
		\end{tikzpicture}
		\ \ \
	}
	\subfigure[NE (\change{$cost=p$})]{
		\ \
		\begin{tikzpicture}[line cap=round,line join=round,x=0.64cm,y=0.64cm]
		\clip(-1.6,0.5) rectangle (1.6,3.53);
		\fill[color=white,fill=white,fill opacity=0] (-0.8,0.8) -- (0.8,0.8) -- (1.2944271909999159,2.3216904260722453) -- (0.,3.2621468297402023) -- (-1.2944271909999157,2.321690426072246) -- cycle;
		\draw [line width=2.pt] (-0.8,0.8)-- (0.8,0.8);
		\draw [color=white] (0.8,0.8)-- (1.2944271909999159,2.3216904260722453);
		\draw [color=white] (1.2944271909999159,2.3216904260722453)-- (0.,3.2621468297402023);
		\draw [color=white] (0.,3.2621468297402023)-- (-1.2944271909999157,2.321690426072246);
		\draw [color=white] (-1.2944271909999157,2.321690426072246)-- (-0.8,0.8);
		\draw [line width=2.pt] (-1.2944271909999157,2.321690426072246)-- (0.,1.90110553638);
		\draw [line width=2.pt] (0.,3.2621468297402023)-- (0.,1.90110553638);
		\draw [line width=2.pt] (0.,1.90110553638)-- (1.2944271909999159,2.3216904260722453);
		\draw [line width=2.pt] (0.,1.90110553638)-- (0.8,0.8);
		\draw [line width=2.pt] (0.,1.90110553638)-- (-0.8,0.8);
		\draw [line width=2.pt] (1.2944271909999159,2.3216904260722453)-- (0.8,0.8);
		\draw [line width=2.pt] (-0.8,0.8)-- (0.8,0.8);
		\begin{normalsize}
		\draw [color=black ,fill=white,line width=2.pt] (-0.8,0.8) circle (4pt);
		\draw[color=black,fill=white] (-1.0682536183519724,1.2) node {5};
		\draw [color=black, fill=white,line width=2.pt] (0.8,0.8) circle (4pt);
		\draw[color=black,fill=white] (1.2,1.2) node {6};
		\draw[color=white] (0.6353240941552338,1.912075524355764) node {};
		\draw [color=black,fill=white,line width=2.pt] (1.2944271909999159,2.3216904260722453) circle (4pt);
		\draw[color=black ] (1.2944271909999159,2.8) node {4};
		\draw [color=black ,fill=white,line width=2.pt] (0.,3.2621468297402023) circle (4pt);
		\draw[color=black] (0.45,3.3) node {1};
		\draw [color=black ,fill=white,line width=2.pt] (-1.2944271909999157,2.321690426072246) circle (4pt);
		\draw[color=black] (-1.2944271909999157,2.8) node {2};
		\draw [color=black, fill=black,line width=2.pt] (0.,1.9) circle (4pt);
		\draw[color=black] (0.3,2.4) node {3};
		\draw[-{latex[scale=3.0]},line width=1.pt,dash pattern=on 2pt off 2pt,color=red] (-1.1,2.07)-- (-0.19,1.78);
		\draw[-{latex[scale=3.0]},line width=1.pt,dash pattern=on 2pt off 2pt,color=red] (-0.14,3)-- (-0.14,2.11);
		\draw[-{latex[scale=3.0]},line width=1.pt,dash pattern=on 2pt off 2pt,color=red] (1.03,2.4)-- (0.18,2.11);
		\draw[-{latex[scale=3.0]},line width=1.pt,dash pattern=on 2pt off 2pt,color=red] (0.8,1.05)-- (0.28,1.82);
		\draw[-{latex[scale=3.0]},line width=1.pt,dash pattern=on 2pt off 2pt,color=red] (-0.51,0.91)-- (-0.,1.62);
		\end{normalsize}
		\end{tikzpicture}
		\ \
	}
	\vspace{-3mm}
	\caption{\label{fig:game:cost} Example strategy profiles in an SGG-AC when $k=1$ and $\xi=2$. \change{Arrows indicate who accesses whose products}. (a) is not an NE since node 6 is better off not buying. \change{Between the NEs, (c) leads to a lower social cost than (b).}}
\end{figure}

In this section, we define equilibria in the games described in Section~\ref{sec:game}. Then, we analyze the efficiency of the equilibria in terms of price of anarchy \change{(PoA)} and price of stability \change{(PoS)}.

\subsection{Definition and Existence of Equilibria}
\label{sec:analysis:exist}

We use the ubiquitous concept of Nash equilibrium (NE) as our solution concept. We first formally define it. 

\begin{definition}[Nash Equilibrium] \label{defn:nash} %Given a graph $\SG=(\SV,\SE)$, 
A strategy profile $s=(s_{i},s_{-i})$ is a \emph{Nash equilibrium (NE)} if no node can increase its utility by changing its strategy given the strategies of the other nodes, i.e.,
$$\forall i\in \SV,\ \forall \change{s'_{i}} \in \SSS_{i},\ u_{i}((s_{i}, s_{-i})) \geq u_{i}((\change{s'_{i}}, s_{-i})).$$
\end{definition}
Figures~\ref{fig:game:basic} and \ref{fig:game:cost} give examples of NEs in an SGG and an SGG-AC, respectively, with explanations.
Note that a strategy profile is an NE in an SGG if and only if the set of owners is a \emph{$k$-independent dominating set} \cite{kreuter1997greedily}, i.e., any two owners have distance at least $k+1$ and every node has distance at most $k$ to some owner.
%\comment{(EL) Explain what equilibria are graph-theoretically?}
Theorem~\ref{thm:exist} states that an NE always exists in both games.
\begin{theorem}[Existence of Nash Equilibria]\label{thm:exist} An NE exists in any SGG and SGG-AC.
\end{theorem}
	\noindent\textit{Proof}.
	Given $\SG = (\SV, \SE)$, $k$ and $\xi$, the following procedure gives an NE $s$ for any SGG and $s'$ for any SGG-AC. 
	Choose an arbitrary node $i$ in the graph, let $i$ buy a good \change{($s_i = 1$, $s'_i = i$)}, and 
	for each node $j$ within $k$ hops from $i$, let $j$ follow $i$  \change{($s_j = 0$, $s'_j = i)$}.
	Delete $i$ and all nodes within $k$ hops from $i$, and repeat until there is no node left in $\SG$. 
	At the end, every node either buys a good or accesses its $k$-hop neighbor's.

	\change{Each node that accesses its $k$-hop neighbor's good} cannot increase its utility by buying a good since the utility of \change{accessing its $k$-hop neighbor's} is greater than that of buying (and no one follows).
	Each node $i$ that buys a good also cannot increase its utility by following another node because the procedure ensures that there is no node that buys a good and is within distance $k$ from $i$. 
	Therefore, $s$ and $s'$ are NEs for the given SGG and SGG-AC, respectively. \qed

\subsection{Social Inefficiency in Equilibria}
\label{sec:analysis:inefficiency}

%We define the price of anarchy (PoA) and the price of stability (PoS). Then, we use them to analyze efficiency of equilibria.

We now turn to the analysis of NEs in our games. It is important to note that a node that does not access any good can increase its utility by buying a good, without decreasing the utilities of the others in both \change{of} our games (see Tables~\ref{tab:utility} and \ref{tab:utility:cost}).
Thus, if we let $\ST$ be the set of strategy profiles where every node accesses a good and thus gets benefit $b$ \footnote{\change{In SGG, $s\in\ST \Leftrightarrow \forall i\in \SV, \sum_{j\in \kneighbor}s_{j}\geq 1$.}\\
	\change{In SGG-AC, $s\in\ST \Leftrightarrow \forall i\in \SV$, ($s_{i}=i$ or ($s_{i}=j$ and $s_{j}=j$)).}}, then all NEs belong to $\ST$.
Due to the same reason, every socially optimal strategy profile \change{(i.e., strategy profile $s$ maximizing social welfare $\sum_{i\in\SV}u_{i}(s)$)} belongs to $\ST$.
Therefore, to define PoA and PoS, we only need to consider the strategy profiles in $\ST$.
Since all strategy profiles in $\ST$ have the same sum of benefits, we can compare them simply by their \emph{social cost}, which is proportional to the number of nodes buying a good (see Definition~\ref{defn:cost}). Importantly, access costs in SGG-AC cancel out (they are paid by some players to others) and do not affect the social cost.

\begin{definition}[Social Cost] \label{defn:cost} 
Given a graph $\SG=(\SV,\SE)$,
the \emph{social cost} of a strategy profile $s=(s_{1},...,s_{n})\in \ST$ is the sum of prices paid by the nodes, i.e.,
\[ 
cost(s)=
\begin{cases}
	\ p\cdot|\{i\in \SV: s_{i}=1\}| & \text{in SGG} \\
	\ p\cdot|\{i\in \SV: s_{i}=i\}|  & \text{in SGG-AC.} \\
\end{cases}
\]
\end{definition}

\begin{table}[t]
	\vspace{-3mm}
	\small
	\centering
	\caption{\label{tab:analysis}
		Summary of our analysis of efficiency of equilibria.
	}
	{
		\begin{tabular}{c|c|c|c}
			\toprule
			& & in SGG & in SGG-AC \\
			\midrule
			\multirow{2}{*}{$k=1$} & $\POANE$  & \multirow{2}{*}{$\Theta(n)$} & $\Theta(n)$\\
			& $\POSNE$ &  & $\Theta(\xi)$ 
			({$\mathbf{=1}$ if $\xi\leq 2$})\\
			\midrule
			\multirow{2}{*}{$k>1$} & $\POANE$  & \multirow{2}{*}{$\Theta(n/k)$} & $\Theta(\max(n/k, n/\xi))$\\
			& $\POSNE$ & & $\Theta(\xi/k)$ 
			($\mathbf{=1}$ if $\xi\leq 2\lfloor k/2 \rfloor +1$)\\
			\bottomrule
		\end{tabular}
	}
\end{table}

The \emph{price of anarchy (PoA)} is defined as the social cost of the \emph{worst} NE divided by minimum social cost (see Definition~\ref{defn:poa}) and the \emph{price of stability (PoS)} is defined as the social cost of the \emph{best} NE divided by minimum social cost (see Definition~\ref{defn:pos}).
Large PoA and PoS indicate that NEs are socially inefficient.

\begin{definition}[Price of Anarchy] \label{defn:poa} 
	Given a graph $\SG=(\SV,\SE)$ with $n$ nodes, 
	the \emph{price of anarchy (PoA)} is defined as%the \change{ratio of the social cost of the worst NE and the minimum social cost}, i.e.,
	$$\POANE=\frac{\max_{s \in \ST:\ s \text{ is an NE}}cost(s)}{\min_{s \in \ST}cost(s)}.$$
\end{definition}

\begin{definition}[Price of Stability] \label{defn:pos} 
	Given a graph $\SG=(\SV,\SE)$ with $n$ nodes,
	the \emph{price of stability (PoS)} is defined as % the \change{ratio of the social cost of the best NE and the minimum social cost}, i.e.,
	$$\POSNE=\frac{\min_{s \in \ST:\ s \text{ is an NE}}cost(s)}{\min_{s \in \ST}cost(s)}.$$
\end{definition}

\begin{figure}[t]
	\centering
	\vspace{-3mm}
	\begin{tikzpicture}[line cap=round,line join=round,x=0.70cm,y=0.70cm]
	\clip(-1.63,4.5) rectangle (5.7,5.4);
	\draw (0.7,5.4) node[anchor=north west] {{Not Buy}};
	\draw (-1.2,5.4) node[anchor=north west] {{Buy}};
	\draw (3.95,5.4) node[anchor=north west] {{Access}};
	\draw [color=black, fill=black,line width=2.pt] (-1.34906,5.1) circle (4pt);
	\draw [color=black, fill=white,line width=2.pt] (0.54172,5.1) circle (4pt);
	\draw[-{latex[scale=3.0]},line width=1.pt,dash pattern=on 2pt off 2pt,color=red] (3.15,5.1) -- (4,5.1);
	\end{tikzpicture}\\
	\vspace{-2mm}
	\subfigure[Best NE in SGG]{\label{fig:inefficiency:basic}
		\begin{tikzpicture}[line cap=round,line join=round,x=0.625cm,y=0.625cm]
		\clip(4.3,0.8) rectangle (10.8,3.4);
		\draw [line width=2.pt] (6.2,3.)-- (7.1,2.);
		\draw [line width=2.pt] (6.2,2.3)-- (7.1,2.);
		\draw [line width=2.pt] (6.2,1.2)-- (7.1,2.);
		\draw [line width=2.pt] (7.1,2.)-- (7.9,2.);
		\draw [line width=2.pt] (7.9,2.)-- (8.8,3.);
		\draw [line width=2.pt] (7.9,2.)-- (8.8,2.3);
		\draw [line width=2.pt] (7.9,2.)-- (8.8,1.2);
		\draw (5.95,2.05) node[anchor=north west] {.};
		\draw (5.95,1.92) node[anchor=north west] {.};
		\draw (5.95,1.79) node[anchor=north west] {.};
		\draw (8.6,2.05) node[anchor=north west] {.};
		\draw (8.6,1.92) node[anchor=north west] {.};
		\draw (8.6,1.79) node[anchor=north west] {.};
		\draw (4.8,3.2) node[anchor=north west] {\HUGE $\{$};
		\draw (8.9,3.2) node[anchor=north west] {\HUGE $\}$};
		\draw (4,2.45) node[anchor=north west] {$\frac{n}{2}$-$1$};
		\draw (9.8,2.45) node[anchor=north west] {$\frac{n}{2}$-$1$};
		%\draw[-{latex[scale=3.0]},line width=1.pt,dash pattern=on 2pt off 2pt,color=red] (6.52,2.96)-- (7.1,2.3);
		%\draw[-{latex[scale=3.0]},line width=1.pt,dash pattern=on 2pt off 2pt,color=red] (6.3,2.06)-- (6.89,1.9);
	%	\draw[-{latex[scale=3.0]},line width=1.pt,dash pattern=on 2pt off 2pt,color=red] (6.52,1.2)-- (7.1,1.76);
	%	\draw[-{latex[scale=3.0]},line width=1.pt,dash pattern=on 2pt off 2pt,color=red] (7.9,2.3)--(8.52,2.96);
		\draw [color=black, fill=white, line width=2.pt] (6.2,1.2) circle (3.5pt);
		\draw [color=black, fill=white, line width=2.pt] (6.2,2.3) circle (3.5pt);
		\draw [color=black, fill=white, line width=2.pt] (6.2,3.) circle (3.5pt);
		\draw [color=black, fill=black, line width=2.pt] (7.1,2.) circle (3.5pt);
		\draw [color=black, fill=white, line width=2.pt] (7.9,2.) circle (3.5pt);
		\draw [color=black, fill=black, line width=2.pt] (8.8,3.) circle (3.5pt);
		\draw [color=black, fill=black, line width=2.pt] (8.8,2.3) circle (3.5pt);
		\draw [color=black, fill=black, line width=2.pt] (8.8,1.2) circle (3.5pt);
		\end{tikzpicture} 
	}
	\subfigure[Best NE in SGG-AC]{\label{fig:inefficiency:cost}
		\begin{tikzpicture}[line cap=round,line join=round,x=0.625cm,y=0.625cm]
		\clip(4.3,0.8) rectangle (10.8,3.4);
		\draw [line width=2.pt] (6.2,3.)-- (7.1,2.);
		\draw [line width=2.pt] (6.2,2.3)-- (7.1,2.);
		\draw [line width=2.pt] (6.2,1.2)-- (7.1,2.);
		\draw [line width=2.pt] (7.1,2.)-- (7.9,2.);
		\draw [line width=2.pt] (7.9,2.)-- (8.8,3.);
		\draw [line width=2.pt] (7.9,2.)-- (8.8,2.3);
		\draw [line width=2.pt] (7.9,2.)-- (8.8,1.2);
		\draw (5.95,2.05) node[anchor=north west] {.};
		\draw (5.95,1.92) node[anchor=north west] {.};
		\draw (5.95,1.79) node[anchor=north west] {.};
		\draw (8.6,2.05) node[anchor=north west] {.};
		\draw (8.6,1.92) node[anchor=north west] {.};
		\draw (8.6,1.79) node[anchor=north west] {.};
		\draw (4.8,3.2) node[anchor=north west] {\HUGE $\{$};
		\draw (8.9,3.2) node[anchor=north west] {\HUGE $\}$};
		\draw (4,2.45) node[anchor=north west] {$\frac{n}{2}$-$1$};
		\draw[-{latex[scale=3.0]},line width=1.pt,dash pattern=on 2pt off 2pt,color=red] (6.52,2.96)-- (7.1,2.3);
		\draw[-{latex[scale=3.0]},line width=1.pt,dash pattern=on 2pt off 2pt,color=red] (6.3,2.06)-- (6.89,1.9);
		\draw[-{latex[scale=3.0]},line width=1.pt,dash pattern=on 2pt off 2pt,color=red] (6.52,1.2)-- (7.1,1.76);
		\draw[-{latex[scale=3.0]},line width=1.pt,dash pattern=on 2pt off 2pt,color=red] (8.52,2.96)-- (7.9,2.3);
		\draw[-{latex[scale=3.0]},line width=1.pt,dash pattern=on 2pt off 2pt,color=red] (8.74,2.06)-- (8.15,1.9);
		\draw[-{latex[scale=3.0]},line width=1.pt,dash pattern=on 2pt off 2pt,color=red] (8.52,1.2)-- (7.9,1.76);
		\draw (9.8,2.45) node[anchor=north west] {$\frac{n}{2}$-$1$};
		\draw [color=black, fill=white, line width=2.pt] (6.2,1.2) circle (3.5pt);
		\draw [color=black, fill=white, line width=2.pt] (6.2,2.3) circle (3.5pt);
		\draw [color=black, fill=white, line width=2.pt] (6.2,3.) circle (3.5pt);
		\draw [color=black, fill=black, line width=2.pt] (7.1,2.) circle (3.5pt);
		\draw [color=black, fill=black, line width=2.pt] (7.9,2.) circle (3.5pt);
		\draw [color=black, fill=white, line width=2.pt] (8.8,3.) circle (3.5pt);
		\draw [color=black, fill=white, line width=2.pt] (8.8,2.3) circle (3.5pt);
		\draw [color=black, fill=white, line width=2.pt] (8.8,1.2) circle (3.5pt);
		\end{tikzpicture}
	}
	\vspace{-3mm}
	\caption{\label{fig:inefficiency} An example of social inefficiency in an SGG. Assume $k=1$. 
	\change{Arrows indicate who accesses whose products.}
	In this example graph, the best NE in an SGG is (a), whose social cost is \change{$np/2$}.
	In an SGG-AC (with $\xi\leq n/2-1$), however, the best NE is (b), whose social cost is \change{$2p$}, equal to the minimum social cost.}
\end{figure}
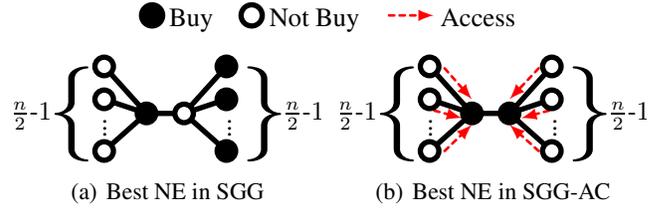

%We show (1) PoA in NE and SNE can be very high without access costs and (2) PoA becomes lower with access costs, by the following two theorems:

In Table~\ref{tab:analysis}, we summarize the results of our worst-case efficiency analysis.
That is, we analyze the two measures in the worst case over all graphs. As stated in Theorem~\ref{thm:poa:basic}, both the PoA and PoS in SGGs are $\Theta(n/k)$ in the worst case.
That is, not only worst NEs but also best NEs can be severely inefficient, with social cost as high as $n/k$ times the optimum.
Figure~\ref{fig:inefficiency:basic} shows an example of such inefficiency for $k=1$, where even the best NEs in the SGG have social cost \change{$np/2$}, while the minimum social cost, in Figure~\ref{fig:inefficiency:cost}, is \change{$2p$}.

\begin{theorem} [PoA and PoS in SGG]
$\POANE$ and $\POSNE$ in SGG are both $\Theta(n/k)$ in the worst case.
\label{thm:poa:basic} 
\end{theorem}
\noindent\textit{Proof}.
For the upper bound \change{on $\POANE$}, fix an arbitrary NE $s \in \ST$. 
Let $k' = \lfloor k / 2 \rfloor$, and for each node $i \in \SV$ that buys a good, consider $\SN_i(k')$, the set of nodes
within distance $k'$ from $i$ (called a {\em ball} around $i$). 
Since each pair of nodes that buy a good are at distance at least $k + 1$ from each other, 
these balls are pairwise disjoint. 

Call a ball $\SN{(k')}$ {\em big} if it has at least $k'$ nodes, and {\em small} otherwise. 
If a ball $\SN_i{(k')}$ is small, $i$ is in a connected component with less than $k'$ nodes, and there is no other node that buys a good in that component. 
Let $c$ be the number of connected components in $\SG$; there are at most $n/k'$ big balls and $c$ small balls. The number of nodes that buy a good is equal to the number of balls, which is at most $n/k' + c$. 

Since the optimal social cost is at least \change{$cp$}, the ratio between the social cost of $s$ and the optimum is at most \change{$(n/k' + c)p/(cp) \leq n/k' + 1 = O(n/k)$}.

For the lower bound \change{on $\POSNE$}, given integers $k, m \geq 1$, 
consider a tree $\SG = (\SV, \SE)$ where there are two center nodes $1$ and $2$, and $2m$ simple paths with $k$ nodes. 
For $m$ of them (called {\em left arms}), one of two endpoints is connected to $1$. 
For the other $m$ of them (called {\em right arms}), one of two endpoints is connected to $2$. 
Finally, $1$ and $2$ are connected. 
\change{Figure~\ref{fig:inefficiency} shows such a graph with $k=1$ and $m=n/2-1$.}
It is easy to see that the optimal social cost is at most $2p$, since if $1$ and $2$ buy a good, all nodes can access at least one good.

We claim that any NE has social cost at least $mp$. Fix an NE $s$. 
In $s$, either $1$ or $2$ does not buy a good, since they can access each other's goods. 
Without loss of generality, suppose that $1$ does not buy a good. 
Consider a left arm, specifically the endpoint of the arm not connected to $1$. 
The only nodes within distance $k$ from this endpoint are $1$ and the nodes in the same arm. 
Since $1$ does not buy a good, there must be a node in the same arm who buys a good. 
This argument holds for each left arm, so at least $m$ nodes buy a good, establishing the claim.

Since the optimal social cost is at most $2p$ and any NE has social cost at least $mp$, 
$\POSNE \geq mp/(2p) = \Omega(n/k)$. 
Since $\POSNE \leq \POANE$, the theorem holds. \qed

Intuitively speaking, the main reason for the inefficiency of NEs in SGGs is that high-degree nodes (i.e., nodes with many $k$-hop neighbors) are less likely to buy
goods even when many neighbors can benefit from goods bought by high-degree nodes.
Indeed, high-degree nodes are more likely to have a neighbor buying a good, and thus choose to free ride.

To incentivize high-degree nodes to buy goods, we can force their neighbors who access the good to pay an access fee to the node --- as we do in SGG-ACs.
In Figure~\ref{fig:inefficiency:cost}, for example, the two high-degree nodes in the center still buy goods, even when they can free ride, since they receive access fees from $\xi$ or more followers, minimizing social cost.

This improvement through access costs is formalized and generalized in Theorem~\ref{thm:pos:cost}, where we show that the worst-case PoS in SGG-ACs is $\Theta(\xi/k)$, which is significantly smaller than $\Theta(n/k)$ in SGGs.
In particular, if $\xi\leq \max(2\lfloor k/2 \rfloor +1, 2)$, then the PoS in SGG-ACs is $1$, i.e., the social cost in the best equilibria is always optimal even in the worst case.
Among $\xi$ values satisfying the condition, the largest one (i.e., $\xi=\max(2\lfloor k/2 \rfloor +1, 2)$) is preferred to minimize the PoA, which is inversely proportional to $\xi$, as shown by Theorem~\ref{thm:poa:cost}.

%Note that this condition holds with larger $\xi$ (i.e., smaller $a$) as $k$ increases.

%Although PoS is improved, the worst-case PoA is still high in SGG-AC, as formalized in Theorem~\ref{thm:poa:cost}.
%However, our simulation in the following section shows that the social costs of realistic NEs obtained by best-response dynamics are close to those of best NEs rather than worst NEs.
\begin{theorem}[PoS in SGG-AC] \label{thm:pos:cost} 
	$\POSNE$ in SGG-ACs is $\Theta(\xi/k)$ in the worst case. In particular, it is $1$ (i.e., there are guaranteed to be socially optimal equilibria) if $\xi \leq \max(2\lfloor k/2 \rfloor + 1, 2$).
\end{theorem}
\noindent\textit{Proof}.
	For the upper bound, given $\SG = (\SV, \SE)$, $k$, and $\xi$, let $\SA^{*} \subseteq \SV$ be a smallest set of the nodes such that for each node $i \in \SV$, there is a node $j \in \SA^{*}$ such that $i$ and $j$ are within distance $k$ from each other (i.e., $|\SA^{*}|\cdot p$ is the optimal social cost).
	Consider a strategy profile $s$ where each node in $\SA^{*}$ buys a good, and
	all other nodes access a good of a node in $\SA^{*}$ such that for each $i \in \SA^{*}$, the set of nodes that access $i$'s good induces a connected subgraph. % of $\SG$. 
	Call a node an {\em owner} if it buys a good.
	An owner $i$ is called a {\em rich owner} if $|\SF_i| \geq \xi$, and a {\em poor owner} otherwise. 
	That is, an owner $i$ is a rich owner if and only if $u_{i}(s)$ is at least the utility of \change{renting a good}. 
	Note that $s\in \ST$ is not an NE if and only if there is a poor owner who can access a good of another owner.
%	In particular, rich owners and free riders do not have any incentive to deviate from the current strategy. 

	From $s$, we show how to construct an NE whose social cost is at most $\Theta(\xi/k) \cdot |\SA^{*}|\cdot p$. 
	If there is a poor owner $i$ who can follow another owner $j$, let $i$ follow $j$. 
	For each node who previously followed $i$, if it can follow another owner, let it follow that owner.
	Call a node {\em  underprivileged} if it previously followed $i$ but cannot follow any other owner. 
	Scan the list of underprivileged nodes sequentially. When $\ell$ is considered, let $\ell$ be an owner (call it a {\em new owner}) and be followed by all still underprivileged nodes who can follow $\ell$ (and remove them from the underprivileged nodes list). 
	At the end of this loop no node is left underprivileged. Repeat until no poor owner $i$ can follow another owner $j$. 

	One of the invariants of this procedure is that any new owner $\ell$ can never access a good of any other owner --- $\ell$ became a new owner since it could not access a good of any other owner, and, subsequently, any node $i\in \SN_\ell(k)$ cannot become a new owner, as it can follow $\ell$, i.e., $\ell\in \SN_i(k)$. Therefore, this procedure \change{always} terminates, after having at most $|\SA^{*}|$ owners follow other owners. 
	The final strategy after termination is an NE since there is no underprivileged node and no poor owner who can follow another owner.
	
To bound the number of new owners, note that when an owner $i \in \SA^{*}$ deviated to follow another owner $j$, 
	among $i$'s previous followers including $i$ (call them $\SC_i$), at most $\max(1, \lfloor \frac{\xi}{\lfloor k/2 \rfloor + 1}\rfloor)$ new owners can be created. 
	This is because $|\SC_i| \leq \xi$ (since $i$ was a poor owner) and if we let $k' = \lfloor k/2 \rfloor$, and consider the ball $\SN_\ell(k')$ around each new owner $\ell$, 
	$|\SN_\ell{(k')} \cap \SC_i| \geq \min(k' + 1, |\SC_i|)$ and all balls are pairwise disjoint (all new owners are at distance at least $k+1$ from each other). 
	The deviation of $i$ creates at most $\max(1, \lfloor \frac{\xi}{k' + 1}\rfloor)$ new owners.
	Therefore, the number of owners in the final NE is at most $|\SA^{*}| \cdot \max(1, \lfloor \frac{\xi}{k' + 1} \rfloor) = O( \frac{\xi}{k} \cdot |\SA^{*}| )$.  
	If $\xi \leq 2 k' + 1$, $\max(1, \lfloor \frac{\xi}{k'+1} \rfloor) = 1$ so the resulting NE is a social optimum. The same conclusion holds when $\xi \leq 2$ since one deviation creates at most one new owner.  
	
	For the lower bound of $\Omega(\xi / k)$, for any integers $k, \xi$ such that $m = \frac{\xi - 1}{k}$ is an integer, build the same tree as in the proof of Theorem~\ref{thm:poa:basic}. As before, the optimal social cost is $2p$, and the social cost at NEs is at least $mp$. To see why the latter claim holds, 
%	consider a tree $\SG = (\SV, \SE)$ where there are two center nodes $1$ and $2$, and $2m$ simple paths with $k$ nodes. 
%	For $m$ of them (called {\em left arms}), one of two endpoints is connected to $1$. 
%	For the other $m$ of them (called {\em right arms}), one of two endpoints is connected to $2$. 
%	Finally, $1$ and $2$ are connected. 
%	It is easy to see that the optimal social cost is at most $2$, since if $1$ and $2$ buy a good, all nodes can access at least one good. 
%	We claim that any Nash equilibrium has social cost at least $m$. Fix any Nash equilibrium $s$. 
%	In $s$, 
note that (as before) $1$ and $2$ cannot simultaneously be owners because the total number of nodes other than $1$ and $2$ is $2(\xi - 1)$, 
	so that at least one of $1$ and $2$ must be a poor owner. Using the same argument as before, 
%	Without loss of generality, let $1$ be not an owner. Then for each left arm, in order for the endpoint not connected to $1$ to access some good, 
%	there must be at least one owner in the same arm. Therefore 
the number of owners is at least $m = \frac{\xi - 1}{k}$, and $\POSNE \geq \frac{\xi - 1}{2k} = \Omega(\frac{\xi}{k})$. \qed

\begin{theorem} [PoA in SGG-AC]
	$\POANE$ in SGG-ACs is $\Theta(\max$ $(n/k,n/\xi))$ in the worst case.
	\label{thm:poa:cost} 
\end{theorem}
\noindent\textit{Proof}.
See Appendix~\ref{sec:appendix:proof}. \qed

%The proof of Theorem~\ref{thm:poa:cost} is relegated to the full version of the paper \cite{shin2017why}.
\change{Note that PoA in SGG-AC has the same order as PoA in SGG if $\xi=\Omega(k)$.}

%\footnote{Note to the reviewers: The formal proof is written up but cannot be provided in any way due to the IJCAI'17 policy. In the camera-ready version of the paper, this footnote will link to the full version.}

%To sum up, the main findings of our equilibrium analysis are as follows:
%\begin{enumerate}
%	\setlength\itemsep{0.01em}
%	\item Inefficiency of NEs in SGG which, however, decreases as $k$ increases (Theorem~\ref{thm:poa:basic})
%	\item Effect of accessing costs on reducing the inefficiency in NEs (Theorem~\ref{thm:pos:cost})
%	\item Proper access costs where $\xi=\lfloor k/2 \rfloor +2$ for minimizing the inefficiency of NEs (Theorems~\ref{thm:pos:cost} and \ref{thm:poa:cost}).
%\end{enumerate}

%\subsection{Computational Complexity Analysis}
%Explain the equivalence of dominating set problem and finding social optima 
%Explain the equivalence of independent dominating set problem  and finding best Nash Equilibria
%Theorem 3 Difficulty of Finding Social Optima
%Theorem 4 Difficulty of Finding Best Nash

\section{Experiments}
\label{sec:simulation}
\change{In this section, we design and conduct experiments to measure social inefficiency in equilibrium, and the effects of $k$ and $\xi$ on the inefficiency. On a high level, our experiments support our qualitative theoretical conclusions.}

\subsection{Algorithms}
\label{sec:simulation:algo}
We use best-response dynamics \change{\cite{matsui1992best}}, described in Algorithm~\ref{alg:best_response}, to find NEs.
A strategy of node $i$ is a \emph{best response} if \change{it} maximizes the utility of $i$ given the others' strategies.
In \emph{best-response dynamics}, nodes iteratively deviate to a best response, until an NE is reached. We run best-response dynamics starting from a strategy profile where no node buys a good.
In both our games, best-response dynamics always converges to an NE, as formalized in Theorem~\ref{thm:converge}.

\begin{algorithm}[h]
	\small
	\caption{\label{alg:best_response} Best-Response Dynamics}
	\SetKwInOut{Input}{Input}
	\SetKwInOut{Output}{Output}
	\Input{
		$\mathcal{G}(\mathcal{V},\mathcal{E})$: network, \\ 
		\ $b$: benefit, $p$: price, $a$: access cost \\
	}
	\Output{
		a strategy profile $s$ corresponding to an NE
	}
	\For{each node $i\in\SV$}{
		\textbf{(in SGG)} $s_{i}\leftarrow 0$ \\
		\textbf{(in SGG-AC)} $s_{i}\leftarrow$ a random strategy in $\kneighbor\backslash\{i\}$
	}
	sort nodes in a random order \label{alg:nash:cost:order}\\
	\While{strategy profile $s$ is not an NE}{
		\For{each node $i\in\SV$ in the sorted order}{
			$u_{max} \leftarrow \max_{x\in \SSS_{i} }u_{i}((x,s_{-i})) $ \\
			\If{$u_{i}((s_{i},s_{-i})) < u_{max} $}{
			$\SB_{i} \leftarrow \{x\in\SSS_{i}:u_{i}(x,s_{-i})=u_{max}\}$\\
			$s_{i}\leftarrow$ a randomly chosen strategy in $\SB_{i}$
			}
		}
	}
	return $s=(s_{1},...,s_{n})$
\end{algorithm}

\begin{theorem} [Convergence of Best-Response Dynamics]\label{thm:converge}
	For both SGGs and SGG-ACs, Algorithm~\ref{alg:best_response} converges to an NE after at most three {\bf while} loops in lines 5--10.
\end{theorem} 

	\noindent\textit{Proof}.
	Assume first that the game is an SGG-AC.
	Given a strategy profile $s$, call a node %{\em privileged} if it is currently accessing a good by either buying a good or free riding, and 
	{\em underprivileged} if it is not accessing a good (it may have neighbors buying a good), and %As before, 
	call a node an {\em owner} if it buys a good.
	There are four types of possible deviations a node can use to improve its utility. 

	%For SGG-AC, an owner $i$ is called a {\em rich owner} if $|\SF_i| \geq \xi$, and a {\em poor owner} otherwise. 
	%In other words, an owner $i$ is a rich owner if and only if $\pi_i(s)$ is at least the utility of free riding. 

%	In SGG, $s$ is a Nash equilibrium if and only if no two owners are within distance $k$ from each other.
%	In SGG-AC, $s$ is a Nash equilibrium if and only if there is no poor owner who can access a good of another node. 

\begin{enumerate}[{Case} 1.]
	\setlength\itemsep{0.01em}
	\item An underprivileged node buys a good since it has no $k$-hop neighbor buying a good.
	\item An underprivileged node \change{rents a good (free rides in SGG)} since it has a $k$-hop neighbor buying a good but has at most $\xi - 1$ followers.
	\item A non-owner buys a good, even though it has a $k$-hop neighbor buying a good, because it has at least $\xi$ followers (who were underprivileged).
	\item An owner changes its action to \change{renting (free riding in SGG)} since it has a $k$-hop neighbor buying a good but has at most $\xi - 1$ followers. 
\end{enumerate}
%	 In SGG, only Case 1, Case 2, and Case 4 are possible, ignoring the conditions on the number of followers for Case 2 and Case 4. 
	In SGGs, only Cases 1 and 4 are possible, ignoring the conditions on the number of followers for Case 4. 
	We now prove that after at most 3 iterations of the {\bf while} loop, $s$ is an NE. 
	\begin{claim}
		After the first iteration, Case 3 cannot happen. \label{claim:1}
	\end{claim}
	\noindent\textit{Proof}.
		Assume for contradiction that Case 3 happened for a node $i$ after the first iteration.
		Right before this, $i$ did not own a good, so either $i$ never owned a good, or it owned a good (via Case 1 or 3) and \change{rented a good} (free rode in SGG) later (via Case 4). 
		Note that a node cannot get any more followers while it does not own a good.
		If $i$ never owned a good, it means all its followers followed it from the beginning, so $i$ should have bought a good in the first iteration, leading to contradiction.
		If $i$ once owned a good but decided to \change{rent a good} (free ride in SGG) later via Case 4, the number of its followers at the moment of the decision was at most $\xi - 1$, and it could not have gained additional followers since then --- again leading to a contradiction. \qed

	\begin{claim}
		After the second iteration, Case 4 cannot happen.
		\label{claim:2}
	\end{claim}
	\noindent\textit{Proof}.
		Assume for contradiction that Case 4 happened for a node $i$ after the second iteration, which means that $i$ gave up its good and decided to \change{rent a good} (free ride in SGG).
		Note Case 1 does not apply to $i$'s $k$-hop neighbors when $i$ owned the item,
		and by Claim~\ref{claim:1}, even Case 3 does not apply after the first iteration. 
		This implies after the first iteration, no new owner appeared among $i$'s $k$-hop neighbors, and the number of $i$'s followers only increased. 
		Thus, either Case 4 should have applied to $i$ earlier, or $i$ should not have bought the good (Case 3 did not apply), leading to a contradiction.  \qed

	\begin{claim}
		After the third iteration, Case 1 and Case 2 cannot happen.
	\end{claim}
	\noindent\textit{Proof}.
		If a node is underprivileged after the third iteration, Case 4 must have happened in the third iteration or later, contradicting Claim~\ref{claim:2}. \qed

	Combining all the claims, $s$ converges to an NE after at most three {\bf while} loops. \qed

Calculating the optimal social cost is also required to measure the efficiency of NEs.
This problem is NP-hard, since it is equivalent to the \emph{minimum \change{$k$-dominating set} problem}, which is known to be NP-hard \cite{hedetniemi1991topics}. Fortunately, we can easily formulate the problem as an integer program, and solve it using intlinprog in MATLAB.% (version R2015a).
%We therefore formulate the problem of finding the optimal social cost as the
%following integer programming problem:
%\begin{align}
%\text{\textbf{minimize}} & \ \ p\sum\nolimits_{i=1}^n s_i\nonumber\\
%\text{\textbf{subject to}} & \ \sum\nolimits_{j\in \kneighbor} s_{j} \geq 1, \forall i \in \mathcal{V}\nonumber\\
% & \ \ s_i \in \{0,1\}, \forall i \in \mathcal{V}\label{ip} 
%\nonumber.
%\end{align}
%That is, the optimal social cost is obtained by minimizing social cost under the constraint that every node is accessible to at least one good (i.e., the strategy profile is included in $\ST$).

\subsection{Datasets}
We run simulations on the following networks:
\bit
	\setlength\itemsep{0.01em}
\item{Synthetic}:
\bit
	\item \textbf{Star} (100 nodes, 99 edges).
	\item \textbf{Chain} (100 nodes, 99 edges).
	\item \textbf{Random} (50 nodes, 127 edges): an Erd\H{o}s-R\'{e}nyi random graph \change{$G(50,0.1)$ (i.e., $50$ nodes and each edge exists i.i.d.~with probability $0.1$)}.
\eit
\item{Real:}
\bit
	\item \textbf{Karate club} (34 nodes, 78 edges): a friendship network between the members of a karate club at a university~\cite{zachary1977information}.
	\item \textbf{Hamsterster} (1,858 nodes, 12,534 edges): a friendship network between the users of Hamsterster, an online community for hamster owners.%\footnote{\scriptsize \url{http://konect.uni-koblenz.de/networks/petster-friendships-hamster}}
	\item \textbf{Advogato} (5,155 nodes, 51,127 edges): a trust network between the users of Advogato, an online community for programmers~\cite{massa2009bowling}.
\eit
\eit

%\begin{table}[h]
%	\vspace{-3mm}
%	\small
%	\centering
%	\caption{\label{tab:data} Datasets used in our simulations }
%	{
%		\begin{tabular}{c|ccc}
%			\toprule
%			& Name & \# Nodes & \# Edges \\
%			\midrule
%			\multirow{3}{*}{Synthetic} & Star & 100 & 99 \\
%			& Chain & 100 & 99 \\
%			& Random & 50 & 127 \\
%			\midrule
%			\multirow{3}{*}{Real} & Karate club & 34 & 78 \\
%			& Hamsterster & 1,858 & 12,534 \\
%			& Advogato & 5,155 & 51,127 \\
%			\bottomrule
%		\end{tabular}
%	}
%\end{table}

\begin{table*}[t]
	\vspace{-3mm}
	\small
	\centering
	\caption{\label{tab:result} Social costs when $k=1$. % Social costs of social optima, NEs in SGG, and NEs in SGG-AC with different access costs are compared. 
		The numbers in the parentheses indicate \change{the standard deviations}.}
	{
		\begin{tabular}{c|c|c|c|c|c|c|c}
			\toprule
			\multirow{3}{*}{Dataset} & \multicolumn{7}{|c}{Social Cost} \\
			\cline{2-8}
			& \multirow{2}{*}{Optimum Cost} & \multirow{2}{*}{NEs in SGGs}  & \multicolumn{5}{|c}{NEs in SGG-ACs} \\
			\cline{4-8}
			&  & & $\xi=1$ & $\xi=2$ & $\xi=5$ & $\xi=10$ & $\xi=20$ \\		
			%&  Optimum & & $a=p/2+\epsilon$ & $a=p/3+\epsilon$ & $a=p/5+\epsilon$ & $a=p/10+\epsilon$ & $a=p/20+\epsilon$ \\		
			%		\cline{4-8}
			%		& Optimum & in SGG &  \\
			\midrule
			Star & 1 & 97.6 (9.50) & \textbf{1.69 (10.9)} & 3.06 (12.6) & 5.92 (21.4) & 10.2 (28.6) & 20.5 (39.1) \\
			Chain & 34 & 43.5 (1.37) & 45.5 (1.65) & \textbf{43.5 (1.36)} & 43.5 (1.38) & 43.5 (1.35) & 43.5 (1.38) \\
			Random & 11 & 17.0 (1.92) & 18.9 (1.46) & \textbf{15.9 (1.70)} & 17.1 (1.91) & 17.1 (1.90) &17.1 (1.90) \\
			\midrule
			Karate club & 4 & 17.0 (3.31) & 10.2 (1.73) & \textbf{8.46 (3.14)} & 14.3 (4.80) & 17.0 (3.36) & 17.1 (3.26)\\
			Hamsterster & 241 & 970 (23.6) & 526 (12.4) & \textbf{426 (16.3)} & 581 (36.8) & 827 (49.8) & 956 (32.1)\\
			Advogato & 806 & 2643 (27.1) & 1468 (22.5) & \textbf{1196 (30.7)} & 1475 (56.2) & 1887 (82.6) & 2199 (101) \\
			\bottomrule
		\end{tabular}
	}
\end{table*}

\begin{table}[t]
	\centering
	\small
	\caption{\label{tab:result:multi_hop} Social costs when $k>1$.}
	{
		\begin{tabular}{c|c|c|c|c}
			\toprule
			\multirow{2}{*}{Dataset} & \multirow{2}{*}{Outcomes} & \multicolumn{3}{|c}{Social Cost} \\
			\cline{3-5}
			& & $k=2$  & $k=3$ & $k=4$ \\
			%&  & & $a=p/2$ & $a=p/3$ & $a=p/5$ & $a=p/10$ & $a=p/20$ \\		
			%		\cline{4-8}
			%		& Optimum & in SGG &  \\
%			\midrule
%			Star & 1 & \\
%			Chain & 34 & \\
%			Random & 11 & \\
			\midrule
			\multirow{3}{*}{Karate club} & Optimum Cost & 2 & 1 & 1 \\
			& NEs in SGGs &  3.23 & 1.76 & 1.28\\
			& NEs in SGG-ACs & 3.23 & 1.75 & 1.26 \\
			\midrule
			\multirow{3}{*}{Hamsterster} & Optimum Cost & 65 & 32 & 28 \\
			& NEs in SGGs & 188 & 90.7 & 46.3\\
			& NEs in SGG-ACs & 128 & 65.0 & 41.7\\
			\midrule
			\multirow{3}{*}{Advogato} & Optimum Cost & 156 & 69 & 58\\
			& NEs in SGGs & 670 & 283 & 102 \\
			& NEs in SGG-ACs & 377 & 129 & 86.5 \\
			\bottomrule
		\end{tabular}
	}
\end{table}

\subsection{Simulation Results}

In Table~\ref{tab:result}, social costs of social optima, NEs in SGGs, and NEs in SGG-ACs are compared.
For NEs, we report the average social cost of $1,000$ NEs returned by Algorithm~\ref{alg:best_response}.
The price $p$ of a good, which is simply a scale factor, is set to $1$.

{\bf Inefficiency of NEs in SGG.}
As seen in Table~\ref{tab:result},
the largest inefficiency (i.e., social cost in NEs / the optimal cost) $97.6$ is obtained on the star graph.
This is because the efficient NE, where only the center node buys a good, is unlikely to be realized, as the center node loses the incentive to buy a good as soon as any of the others does.
Thus, the inefficient NE, where all nodes except the center node buy a good, is realized with high probability.
On the chain graph and the random graph, however, the inefficiency is only $1.28$ and $1.55$, respectively, since high-degree nodes, which are main source of inefficiency, do not exist.
On real networks, where high-degree nodes exist (albeit not as extreme as the star graph), the inefficiency is between $3.28$ and $4.25$.

{\bf Effects of access costs on reducing the social inefficiency.}
As seen in Table~\ref{tab:result}, the inefficiency significantly decreases in SGG-ACs with an appropriate access cost, compared to SGGs.
In particular, on the star graph, the inefficiency decreases by $93\%$.
This is because the inefficient NE, where all nodes except the center node buy the good, is unlikely to be realized, as the center node can still buy a good even when it has neighbors buying goods.
For similar reasons, the inefficiency on real networks also decreases by $50\%-56\%$.
Since the inefficiency on the chain and random graphs is already low in SGGs, the improvement is smaller.

{\bf Access costs that maximize efficiency.}
As seen in Table~\ref{tab:result}, inefficiency is lowest at $\xi=2$ ($p/3<a<p/2$), consistent with our suggestion of $\xi=\max(2\lfloor k/2 \rfloor +1, 2)$ in Section~\ref{sec:analysis}. \change{The inefficiency tends to increase as $\xi$ increases}.

{\bf Effects of $k$.}
Social costs on real networks when $k>1$ are shown in Table~\ref{tab:result:multi_hop}, where we set $\xi$ to $6$ on Karate club and $4$ on the others.
In both SGGs and SGG-ACs, inefficiency in NEs decreases as $k$ increases, which is consistent with Theorems~\ref{thm:poa:basic}, \ref{thm:pos:cost}, and \ref{thm:poa:cost}.
Although the inefficiency is still smaller in SGG-ACs than in SGGs, the gap decreases as $k$ increases.
%Although not presented in the table, the suggested access costs, where $\xi=\lfloor k/2 \rfloor +2$, lead to social costs close (on average $10\%$ higher) to social costs with the optimal access costs.

\section{Related Work}
\label{sec:related}
%Our model is related to the literature on network games, which has received considerable interest in recent decade.
We first review previous work directly related to network games with shareable goods. See the work of  \citeauthor{galeotti2010network}~\shortcite{galeotti2010network}, and the chapter by Jackson and Zenou~\shortcite{jackson2014games}, for an introduction to network games in general.

\citeauthor{bramoulle2007public}~\shortcite{bramoulle2007public} and \citeauthor{bramoulle2014strategic}~\shortcite{bramoulle2014strategic} study a network game where the strategy of each node is its contribution to a public good, and its utility is a function of its own contribution and that of its direct neighbors.

\citeauthor{ballester2006s}~\shortcite{ballester2006s} consider a similar game where, however, the utility of each node is a concave function of its effort and a linear function of its neighbors'.
With these conditions, the game has a unique NE where the effort of each node is proportional to its Bonacich centrality score.
\citeauthor{elliott2013network}~\shortcite{elliott2013network} study an extended game in directed, weighted graphs where an edge $(i,j)$ indicates the marginal benefit that node $i$ can provide to node $j$.
\citeauthor{allouch2015private}~\shortcite{allouch2015private} develops a model where consumptions of both private goods and public goods are taken into account.
%\cite{bloch2007effect} considers a game where the utility of each player depends on the amount of public goods provided in its community and the amount of spillovers from other communities.
%If spillovers are symmetric, a unique NE exists, and the total amount of goods decreases as spillovers increase. 

In our models (of Section~\ref{sec:game}), each node simply decides whether to buy a good or access its neighbors', rather than the amount of effort.
Instead, we extend the previous models, \change{especially the best-shot game \cite{hirshleifer1983weakest}}, in that (a)~access costs can be imposed on free riders and (b) nodes can benefit not only from direct neighbors but also $k$-hop neighbors.
In contrast to previous work, we analyze PoA and PoS, and provide empirical results on real social networks.
%Moreover, we provide PoA and PoS analysis with simulation on real-world graphs, which 
%We believe that this is the first work that shows %how to optimize a non-submodular function...

Our work is also related to the huge body of literature on the price of anarchy. The concept itself is due to \citeauthor{KP99}~\shortcite{KP99}, and the price of stability was introduced a few years later by \citeauthor{ADKT+04}~\shortcite{ADKT+04}. These concepts underlie much work at the intersection of game theory and AI, e.g., in computational social choice~\cite{BCMP13}, security games~\cite{LV15}, and routing~\cite{VFH15}.

To the best of our knowledge, the price-of-anarchy paper that is most closely related to ours is the one by \citeauthor{KPR13}~\shortcite{KPR13}. They give bounds on the price of anarchy of an anti-coordination game played on a graph, albeit a fundamentally different one: each player chooses a color, and the utility of a player is the number of neighbors with different colors. In their work, $k$-hop neighbors are not considered, and access costs are incompatible with the model.
It is also worth mentioning that our use of access costs to reduce the inefficiency of equilibria is conceptually related to work on taxation in congestion games~\cite{CKK10}.

\section{Final Words}
\label{sec:conclusion}
In our view, the most actionable conclusion from our work is that in the type of scenarios under consideration (shareable goods on a network), access costs should be imposed when possible. Despite the whimsical title of our paper, this would be hard to do at a societal level for things like ski equipment and portable cribs. However, it certainly seems feasible at the level of an organization. For example, a university could mandate access costs for expensive lab equipment bought by individual researchers, as this would actually decrease the amount of grant money that is invested in buying equipment. For the designer of a multi-agent system, imposing access costs is trivial, and, similarly, might lead (\emph{would} lead, if one trusts our analysis) to significant benefits.

%In this work, we propose game-theoretic models that capture incentives to buy a good sharable with $k$-hop neighbors on a social network.
%With our models, we find that social inefficiency, specifically overproduction of goods, can occur in Nash Equilibria.
%However, we also show that this inefficiency can be reduced significantly by sharing goods with more players (i.e., increasing $k$) or imposing access costs to free riders.
%We provide efficiency analysis of equilibria (in terms of PoA and PoS) and simulations using  best-response dynamics on real-world social networks, with thousands of nodes, to support our findings.

\section*{Acknowledgements}
\change{We thank Prof. Christos Faloutsos for fruitful discussions.}


\balance
\bibliographystyle{named}
\bibliography{abb,ultimate,kijung}

\appendix
%\newpage
\section{Proof of Theorem~\ref{thm:poa:cost}}
\label{sec:appendix:proof} 
\noindent\textit{Proof}.
	For the upper bound, fix an arbitrary NE $s \in \ST$. 
	Let $O$ be the set of nodes who buy a good in $s$, and call them {\em owners}. 
	For each owner $i \in O$, we call $i$ a {\em rich owner} if $|\SF_i| \geq \xi$, i.e., $u_i(s)$ is at least the utility of \change{renting a good}. 
	Otherwise, call $i$ a {\em poor owner}. 
	The number of rich owners is at most $n/(\xi + 1)$. 
	
	Note that any two poor owners are at distiance at least $k + 1$ from each other, since otherwise they 
	prefer to access the other's good. 
	Let $k' = \lfloor k / 2 \rfloor$, and for each poor owner $i \in \SV$, consider $\SN_i(k')$, the set of nodes
	within distance $k'$ from $i$ (called a {\em ball} around $i$). 
	Since each pair of poor owners are at distance at least $k + 1$ from each other, these balls are pairwise disjoint. 
	
	Call a ball $\SN_i(k')$ {\em big} if it has at least $k'$ nodes, and {\em small} otherwise. 
	If a ball $\SN_i(k')$ is small, $i$ is in a connected component with less than $k'$ nodes, and there is no other owner in that component. 
	Let $c$ be the number of connected components in $G$. 
	Since there are at most $n/k'$ big balls and $c$ small balls, the number of poor owners is equal to the number of balls, which is at most $n/k' + c$. 
	The total number of owners is at most $n/k' + n/(\xi + 1) + c$. 
	
	Since the optimal social cost is at least \change{$cp$}, the gap between the optimal social cost and the social cost of $s$ is $\frac{(n/k' + n/(\xi + 1) + c)p}{cp} = O(n/k + n/\xi) = O(\max(n/k, n/\xi))$.

	For the lower bound of $\Omega(n/k)$, for any integers $k, m, \xi \geq 1$, 
	consider a tree $\SG = (\SV, \SE)$ where there is a center node $1$, and $m$ simple paths with $k$ nodes. 
	For $m$ of them (called {\em arms}), one of two endpoints is connected to $1$. 
	It is easy to see that the optimal social cost is at most $p$, since if $1$ buys a good, all nodes can access it.
	But if we consider a strategy profie where the endpoint not connected to $1$ buys a good for each arm and each node
	follows the endpoint of its own arm ($1$ follows an arbitrary one), it is an NE of social cost $m$. 
	Therefore, the gap is $m = \Omega(n / k)$. 
	
	For the lower bound of $\Omega(n/\xi)$, for any integers $k, m, \xi \geq 1$, 
	let $\SG = (\SV, \SE)$ be the complete graph on $m(\xi + 1)$ vertices. 
	Obviously \change{the optimal social cost is $p$}, but if $m$ nodes buy a good and have $\xi$ followers each, 
	it becomes an NE \change{with social cost $mp$}. The gap is $\Omega(n/\xi)$. Combining the two lower bounds, the theorem is proved. \qed

\end{document}